\documentclass[10pt,twocolumn,letterpaper]{article}

\usepackage[pagenumbers]{cvpr} 

\usepackage{graphicx}
\usepackage{amsmath}
\usepackage{amssymb}
\usepackage{booktabs}
\usepackage{amsfonts}       
\usepackage{nicefrac}       
\usepackage{kotex}
\usepackage{microtype}      
\usepackage{multirow}
\usepackage{makecell}
\usepackage{pifont}         
\usepackage{enumitem}
\usepackage{scalerel}
\usepackage{soul}
\usepackage{setspace}

\usepackage[pagebackref,breaklinks,colorlinks]{hyperref}

\usepackage[capitalize]{cleveref}
\crefname{section}{Sec.}{Secs.}
\Crefname{section}{Section}{Sections}
\Crefname{table}{Table}{Tables}
\crefname{table}{Tab.}{Tabs.}

\begin{document}

\title{Bridging the Gap between Classification and Localization\\ for Weakly Supervised Object Localization}

\author{Eunji Kim$^1$ ~~~~~~~ Siwon Kim$^1$ ~~~~~~~  Jungbeom Lee$^1$  ~~~~~~~  Hyunwoo Kim$^2$  ~~~~~~~  Sungroh Yoon$^{1, 3}$\thanks{Correspondence to: Sungroh Yoon (sryoon@snu.ac.kr).}\\
$^1$ Department of Electrical and Computer Engineering, Seoul National University ~
$^2$ LG AI Research\\
$^3$ Interdisciplinary Program in AI, AIIS, ASRI, INMC, and ISRC, Seoul National University\\
{\tt\small \{kce407, tuslkkk, jbeom.lee93\}@snu.ac.kr, hwkim@lgresearch.ai, sryoon@snu.ac.kr}}
\maketitle

\newcommand{\xmark}{\text{\ding{55}}}
\newcommand{\cmark}{\text{\ding{51}}}

\begin{abstract}
Weakly supervised object localization aims to find a target object region in a given image with only weak supervision, such as image-level labels. Most existing methods use a class activation map (CAM) to generate a localization map; however, a CAM identifies only the most discriminative parts of a target object rather than the entire object region. In this work, we find the gap between classification and localization in terms of the misalignment of the directions between an input feature and a class-specific weight. We demonstrate that the misalignment suppresses the activation of CAM in areas that are less discriminative but belong to the target object. To bridge the gap, we propose a method to align feature directions with a class-specific weight. The proposed method achieves a state-of-the-art localization performance on the CUB-200-2011 and ImageNet-1K benchmarks.
\end{abstract}

\section{Introduction}
\label{sec:intro}
Object localization aims to find the area of a target object in a given image~\cite{ren2015faster,russakovsky2015imagenet,duan2019centernet,lin2017feature,tan2020efficientdet}. However, fully supervised approaches require accurate bounding box annotations, which require a tremendous cost. Weakly supervised object localization (WSOL) has been a great alternative because it requires only image-level labels to train a localization model~\cite{singh2017hide,choe2019attention,choe2020evaluation,pan2021unveiling, xue2019danet}.

The most commonly used approach for WSOL is a class activation map (CAM)~\cite{zhou2016learning}. CAM-based methods employ a global average pooling (GAP) layer~\cite{lin2013network} followed by a fully connected (FC) layer, and generate a CAM with the feature maps prior to the GAP layer.
A highly activated area in a CAM is predicted to be an object location.
However, it is widely observed that CAM identifies only the most discriminative parts of an object rather than the entire object area, resulting in low localization performance~\cite{mai2020erasing,lee2021anti,zhang2018adversarial}.

\begin{figure}[t]
	\centering
    \includegraphics[width=0.95\columnwidth]{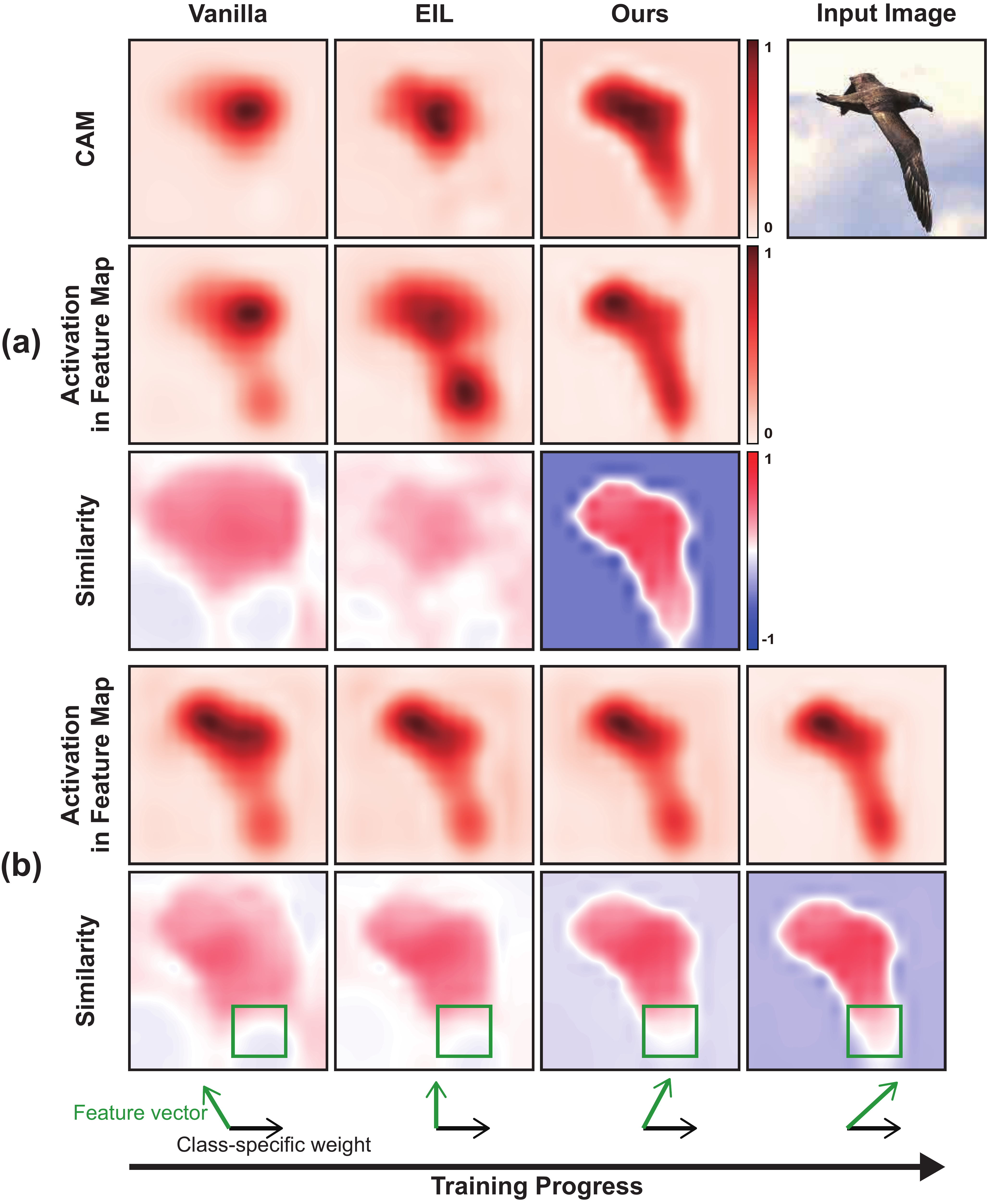}
    \vspace{-0.5em}
    \caption{(a) Examples of CAM and decomposed terms from the classifier trained with the vanilla method~\cite{zhou2016learning} and with EIL~\cite{mai2020erasing}. (b) Visualization of the changes of CAM and decomposed terms as training with our method progresses.}
    \label{fig:first}
\end{figure}

We ask the question, ``\textit{Why does CAM generated from an accurate classifier fail to highlight the entire object area?}''
To answer this, we provide a new perspective of decomposing CAM into two terms: (1) activation in a feature map and (2) cosine similarity between the feature vector at each spatial location and the class-specific weight in the FC layer.
Fig.~\ref{fig:first}(a) shows that only the bird's body is highly activated in the CAM of the vanilla model, leaving the wing less activated. However, looking at the activation in the feature map, the wing as well as the body is highly activated.
The low similarity of the wing region offsets the activation in the feature map, making the region invisible in the CAM.
Here, we find that the low cosine similarity, \ie, misalignment of feature directions to the class-specific weights, prevents the less discriminative part belonging to a target object from being highly activated in a CAM.
This is because training for classification only considers the feature averaged over all locations, not the feature at each spatial location.
This brings the gap between classification and localization.

Although various approaches have been proposed to expand the activated region to the entire object area in a CAM~\cite{zhang2018adversarial,choe2019attention,mai2020erasing,yun2019cutmix,xue2019danet,zhang2018self}, none of them discovered or mitigated the misalignment. Fig.~\ref{fig:first}(a) shows that EIL~\cite{mai2020erasing}, one of those approaches, expands the activated region in the feature map. However, it fails to increase the similarity in the object region; hence, the expansion effect is not as large in the CAM as in the activation of the feature map.

To bridge the gap between classification and localization, we propose feature direction alignment, a method to enhance the alignment of feature directions in the entire object region to the directions of class-specific weights while discouraging the alignment in the background region.
We also introduce consistency with attentive dropout, which ensures that the target object region has uniformly high activation in the feature map.
Fig.~\ref{fig:first}(b) shows that our method gradually aligns the feature directions to the class-specific weight as the training progresses.
The alignment results in high activation of less discriminative regions, \eg, wing, in the CAM, enabling accurate localization of the entire object.
We evaluate our method on the most widely used WSOL benchmark datasets: CUB-200-2011~\cite{welinder2010caltech} and ImageNet-1K~\cite{russakovsky2015imagenet}.
Our method achieves a state-of-the-art localization performance for both datasets.

The contributions of this paper can be summarized as follows:
\begin{itemize}
\setlength{\itemsep}{2pt}
\vspace{-3pt}
	\item[$\bullet$] We interpret a CAM in terms of the degree of alignment between the direction of input features and the direction of class-specific vectors, and find the gap between classification and localization.
	\vspace{-2pt}
	\item[$\bullet$] We propose a method to bridge the gap between classification and localization by aligning feature directions with class-specific weights.
	\vspace{-2pt}
	\item[$\bullet$] We demonstrate that our proposed method outperforms other state-of-the-art WSOL methods on the CUB-200-2011 and ImageNet-1K datasets.
\end{itemize}
\section{Related Work}
The WSOL method trains a model to localize objects using image-level labels. Zhou~\etal~\cite{zhou2016learning} introduce a CAM to identify the location of a target object via GAP layer~\cite{lin2013network}. However, it fails to identify the entire object region.

Various methods have been proposed to activate the entire object region in a CAM. HaS~\cite{singh2017hide} trains a classifier using images that are erased with a random patch. ACoL~\cite{zhang2018adversarial} employs two parallel classifiers to identify complementary regions. ADL~\cite{choe2019attention,choe2020attention} stochastically drops out the attentive feature in a single forward pass. Ki~\etal~\cite{ki2020sample} introduced contrastive learning with foreground features and background features. EIL~\cite{mai2020erasing} adopts an additional forward pass to classify with the feature whose highly activated regions are erased. SPG~\cite{zhang2018self} utilizes a deep feature to guide a shallow feature and $\text{I}^2\text{C}$~\cite{zhang2020inter} uses pixel-level correlations between two different images. CutMix~\cite{yun2019cutmix} combines two patches from different images and assigns a new class label based on the area of each patch. DANet~\cite{xue2019danet} leverages divergent activations with the hierarchy of classification labels.

\begin{figure}[t]
	\centering
    \includegraphics[width=0.95\columnwidth]{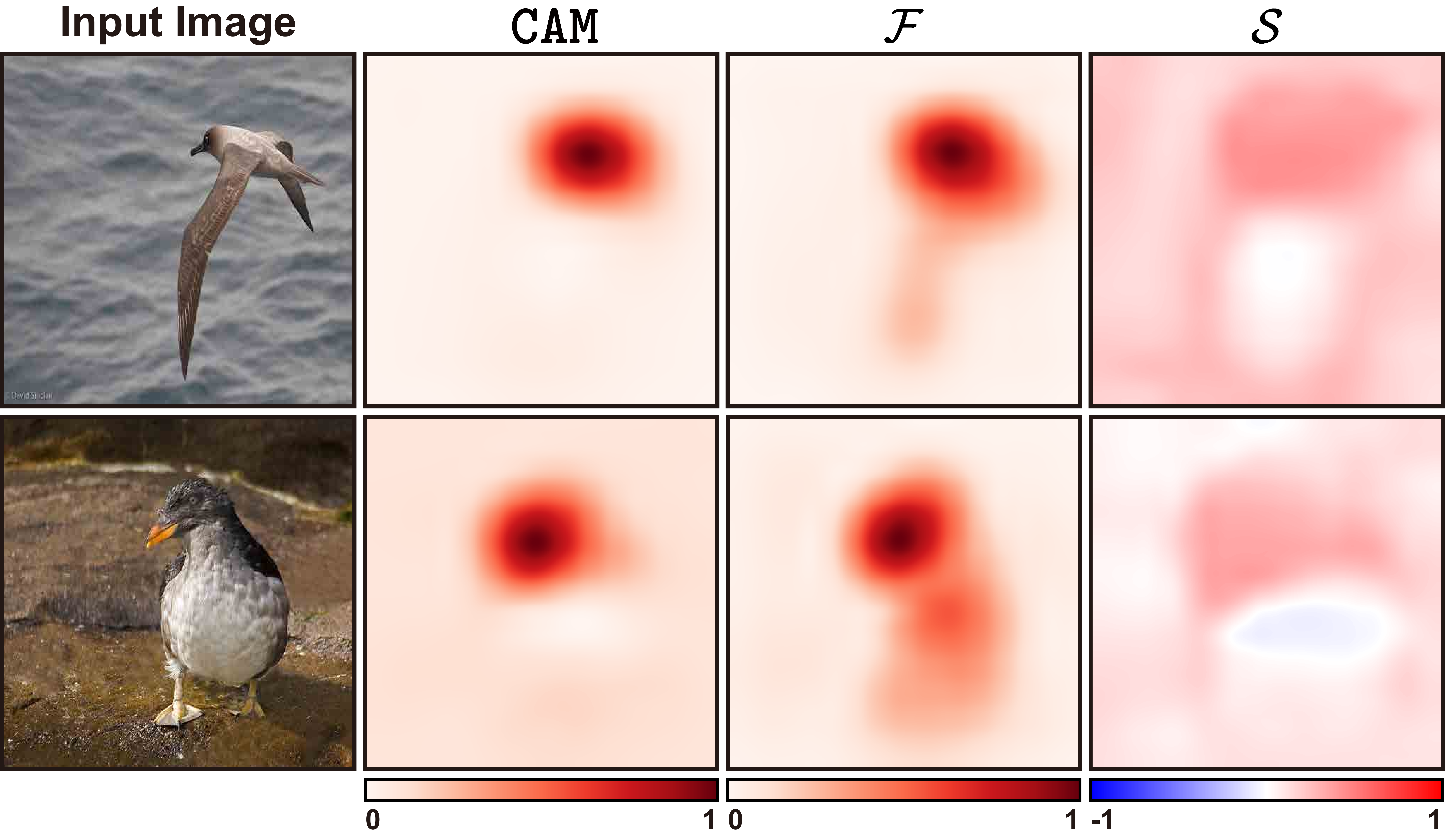}
    \vspace{-0.5em}
    \caption{Examples of CAM and decomposed terms $\mathcal{F}$ and $\mathcal{S}$ from a vanilla model. The CAMs and $\mathcal{F}$ are normalized as in $[0, 1]$ for visualization. It shows the misalignment of the feature directions with the class-specific weights.
    }
    \label{fig:cam_norm_sim}
\end{figure}
\begin{figure*}[t]
	\centering
    \includegraphics[width=0.95\textwidth]{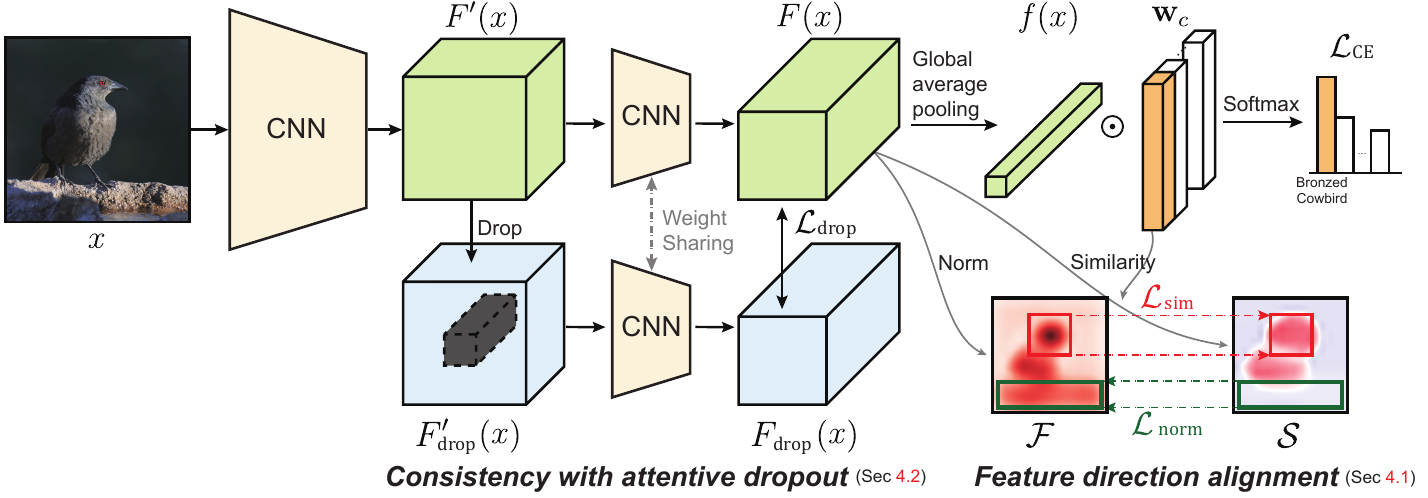}
    \vspace{-0.5em}
    \caption{Overview of the proposed method. It consists of two strategies: feature direction alignment and consistency with attentive dropout.}
    \label{fig:overall}
\end{figure*}

There have been attempts to obtain localization maps in different ways, pointing out the limitations of CAM-based methods. Pan~\etal~\cite{pan2021unveiling} proposed a method to utilize high-order point-wise correlation to generate localization maps. Kim~\etal~\cite{kim2021keep} proposed a CALM that learns to predict the location of the cue for recognition.

Several normalization methods have been proposed to obtain the bounding boxes around predicted object locations from a continuous localization map. Bae~\etal~\cite{bae2020rethinking} proposed several methods to address the bias in GAP, including a new normalization method, PaS, which restricts the maximum value of the activation map. IVR~\cite{kim2021normalization} is a normalization method that restricts the minimum value of the activation map.

Some works have adopted an auxiliary module for localization besides classification. GC-Net~\cite{lu2020geometry} adopts a separate detector for localization trained with a geometric constraint. FAM~\cite{meng2021foreground} generates a class-agnostic foreground map through a memory mechanism. ORNet~\cite{xie2021online} adopts an additional activation map generator and refines the activation map in an online manner. PSOL~\cite{zhang2020rethinking}, SLT-Net~\cite{guo2021strengthen}, and SPOL~\cite{wei2021shallow} use two separate networks for classification and localization.

Our method aims to address the gap between classification and localization without adopting any auxiliary module. The methods that adopt additional modules or even separate models use more parameters and computational resources.  Therefore, we compare our method mainly with the WSOL methods that use a single branch, for a fair comparison.
\section{Finding the Gap with CAM Decomposition}
Given an input image $x$ and a typical image classifier comprising convolutional layers and a GAP followed by an FC layer, a CAM for target class $c$ is computed as follows:
\begin{equation}\label{eq:cam}
\texttt{CAM}(x) = \mathbf{w}^\intercal_c F(x).
\end{equation}
$F(x)\in\mathbb{R}^{H \times W \times D}$ is the feature map before the GAP, and $\mathbf{w}_c\in\mathbb{R}^{D}$ is the weight of the FC layer connected to class $c$, where $H$, $W$, and $D$ are the height, width, and dimension, respectively.
Eq.~\ref{eq:cam} implies that the value of CAM at each spatial location is the dot product of two vectors, $\mathbf{w}_c$ and $F_u(x)$, where $u\in\{1, ..., HW\}$ is the index of spatial location.
It can be decomposed as follows:
\begin{equation}\label{eq:cam_each}
\begin{aligned}
\texttt{CAM}_u(x) = & \mathbf{w}_c \cdot F_u(x) \\
= & \|\mathbf{w}_c\|\|F_u(x)\| \underbrace{\frac{\mathbf{w}_c \cdot F_u(x)}{\|\mathbf{w}_c\|\|F_u(x)\|}}_{\textstyle S(\mathbf{w}_c,F_u(x))},
\end{aligned}
\end{equation}
where $S(\mathbf{a},\mathbf{b})$ is the cosine similarity between the two vectors, $\mathbf{a}$ and $\mathbf{b}$.
When generating a CAM, target class $c$ is fixed and $\|\mathbf{w}_c\|$ is the same for every $u$.
The CAM value at each position can now be interpreted as the product of the norm of the feature vector at the corresponding location and the similarity between the feature vector and class-specific weight vector.
Let $\mathcal{F}\in\mathbb{R}^{H \times W}$ and $\mathcal{S}\in\mathbb{R}^{H \times W}$ be the norm map and the similarity map, respectively, where $\mathcal{F}_u=\|F_u\|$ and $\mathcal{S}_u=S(\mathbf{w}_c, F_u(x))$. Subsequently, CAM can be rewritten as
\begin{equation}\label{eq:cam_abb}
\texttt{CAM}(x) = \|\mathbf{w}_c\|\cdot\mathcal{F}\odot\mathcal{S}.
\end{equation}
To localize the target object accurately, both $\mathcal{F}_u$ and $\mathcal{S}_u$ should be large for $u$ belonging to the object.

Likewise, the classification score can be interpreted with the output of the GAP, $f(x)=\text{GAP}(F(x))\in\mathbb{R}^{\rm{D}}$.
\begin{equation}\label{eq:logit}
\begin{aligned}
\texttt{logit}_c(x) = & \mathbf{w}_c \cdot f(x) \\
= & \|\mathbf{w}_c\|\left\lVert f(x)\right\rVert S\left( \mathbf{w}_c,f(x)\right).
\end{aligned}
\end{equation}
Because $\left\lVert f(x)\right\rVert$ is fixed for $x$, $\|\mathbf{w}_c\|$ and $S\left( \mathbf{w}_c,f(x)\right)$ determine the logit score of each class $c$.
The scale variation of $\|\mathbf{w}_c\|$ across classes is not very large.
Therefore, to classify $x$ correctly, $S(\mathbf{w}_c, f(x))$ must be large for the ground truth class $c$.
Here exists the gap between classification and localization.
The classifier is trained to increase $S(\mathbf{w}_c, f(x))$, not $S(\mathbf{w}_c, F_u(x))$ for $u$ belonging to an object region. Cosine similarity is interpreted as the degree of alignment between the directions of the two vectors, meaning that the input feature vector at the object region and class-specific weight vector are not ensured to be aligned with training only for classification.
This causes the model to fail to localize the entire object in a CAM.

Fig.~\ref{fig:cam_norm_sim} shows some examples of norm map $\mathcal{F}$, similarity map $\mathcal{S}$, and CAM from a vanilla model.
The less discriminative but object-belonging regions also have noticeably high activation in $\mathcal{F}$, including wings and bodies of birds. 
However, those regions are not activated in the final CAMs, due to the small values in $\mathcal{S}$.
Although $\mathcal{F}$ contains considerable information for localization, its effect diminishes because of the misalignment of the feature directions with the class-specific weight.

In the next section, we propose a method to bridge the gap between classification and localization by aligning feature directions: adjusting the cosine similarity between input features and class-specific weights.

\section{Bridging the Gap through Alignment}
We describe how to align feature directions in Sec.~\ref{sec:feature_directions}. An additional strategy to enhance the effect of the feature direction alignment, consistency with attentive dropout, is introduced in Sec.~\ref{sec:consistency_drop}. In Sec.~\ref{sec:training_scheme}, we describe the overall training scheme.
Fig.~\ref{fig:overall} shows the overview of our proposed method.

\subsection{Alignment of Feature Directions}\label{sec:feature_directions}
To enhance the activation of the entire object region in CAM, we want the cosine similarity between $F_u$ and $\mathbf{w}_{c}$ to be high for $u$ belonging to the target object and low for the background region.
Because high activation in $\mathcal{F}$ implies that there is a cue for classification at the corresponding location, we divide the region of the feature map into coarse foreground region $\mathcal{R}^\text{norm}_\text{fg}$ and background region $\mathcal{R}^\text{norm}_\text{bg}$ based on a normalized $\mathcal{F}$.
\begin{equation}
\begin{aligned}
&\mathcal{R}^\text{norm}_\text{fg}=\{u|\hat{\mathcal{F}}_u>\tau_\text{fg}\},\\
&\mathcal{R}^\text{norm}_\text{bg}=\{u|\hat{\mathcal{F}}_u<\tau_\text{bg}\},\\
&\text{where}~\hat{\mathcal{F}}=\frac{\mathcal{F}-\min_{i}{\mathcal{F}_i}}{\max_{i}{\mathcal{F}_i}-\min_{i}{\mathcal{F}_i}}.
\end{aligned}
\end{equation}
$\tau_\text{fg}$ and $\tau_\text{bg}$ are constant thresholds that determine the foreground and background regions, respectively. Note that $\tau_\text{fg}$ and $\tau_\text{bg}$ are not the same; therefore, there is an unknown region that is not included in either $\mathcal{R}^\text{norm}_\text{fg}$ or $\mathcal{R}^\text{norm}_\text{bg}$.
To increase $\mathcal{S}_u$ in $\mathcal{R}^\text{norm}_\text{fg}$ and suppress it in $\mathcal{R}^\text{norm}_\text{bg}$, we define the similarity loss as follows:
\begin{equation}\label{eq:loss_sim}
\begin{aligned}
\mathcal{L}_\text{sim} = -\frac{1}{|\mathcal{R}^\text{norm}_\text{fg}|}\sum_{u\in \mathcal{R}^\text{norm}_\text{fg}}{\mathcal{S}_u} +\frac{1}{|\mathcal{R}^\text{norm}_\text{bg}|}\sum_{u\in \mathcal{R}^\text{norm}_\text{bg}}{\mathcal{S}_u}.
\end{aligned}
\end{equation}

There still remains a possibility that some parts of the object region have low activation in $\hat{\mathcal{F}}$.
In this case, $\mathcal{L}_\text{sim}$ may not be sufficient for the alignment.
Therefore, we introduce an additional loss term to increase $\hat{\mathcal{F}}$ in every candidate region belonging to the target object.
Because a positive $\mathcal{S}_u$ indicates that $u$ is making a positive contribution to increasing the classification logit, the regions with positive similarity can be treated as candidates for the object region. Therefore, we force this area to be activated.
We estimate the object region, $\mathcal{R}^\text{sim}_\text{fg}$, and background region, $\mathcal{R}^\text{sim}_\text{bg}$, based on $\mathcal{S}_u$ as
\begin{equation}
\begin{aligned}
&\mathcal{R}^\text{sim}_\text{fg}=\{u|\mathcal{S}_u>0\},\\
&\mathcal{R}^\text{sim}_\text{bg}=\{u|\mathcal{S}_u<0\}.
\end{aligned}
\end{equation}
With each estimated region, we define the norm loss in a manner similar to Eq.~\ref{eq:loss_sim}, as follows:
\begin{equation}\label{eq:loss_norm}
\begin{aligned}
\mathcal{L}_\text{norm} = -\frac{1}{|\mathcal{\mathcal{R}^\text{sim}_\text{fg}}|}\sum_{u\in \mathcal{R}^\text{sim}_\text{fg}}{\hat{\mathcal{F}}_u} +\frac{1}{|\mathcal{R}^\text{sim}_\text{bg}|}\sum_{u\in \mathcal{R}^\text{sim}_\text{bg}}{\hat{\mathcal{F}}_u}.
\end{aligned}
\end{equation}

For fine-grained classification, such as bird species classification, the object to be recognized is the same across classes. In this case, we define the region with a non-positive similarity with any class as $\mathcal{R}^\text{sim}_\text{bg}$ and the other as $\mathcal{R}^\text{sim}_\text{fg}$. In general, the regions $\mathcal{R}^\text{sim}_\text{bg}$ and $\mathcal{R}^\text{sim}_\text{fg}$ are defined with a similarity with a target class.

The two loss terms $\mathcal{L}_\text{sim}$ and $\mathcal{L}_\text{norm}$ operate complementary.
Through the minimization of $\mathcal{L}_\text{sim}$, the value of $\mathcal{S}$ in the region that is highly activated in $\hat{\mathcal{F}}$ increases.
Through the minimization of $\mathcal{L}_\text{norm}$, the value of $\hat{\mathcal{F}}$ in the region with high similarity increases.
After the joint minimization of $\mathcal{L}_\text{sim}$ and $\mathcal{L}_\text{norm}$, the activated region in $\hat{\mathcal{F}}$  and that in $\mathcal{S}$ become similar.

\subsection{Consistency with Attentive Dropout}\label{sec:consistency_drop}
We can expect the successful alignment by $\mathcal{L}_\text{sim}$ when the estimation of $\mathcal{R}^\text{norm}_\text{fg}$ and $\mathcal{R}^\text{norm}_\text{bg}$ is accurate: $\hat{\mathcal{F}}$ is consistently large over the entire object region and small over the background region.
\begin{figure}[t]
	\centering
    \includegraphics[width=\columnwidth]{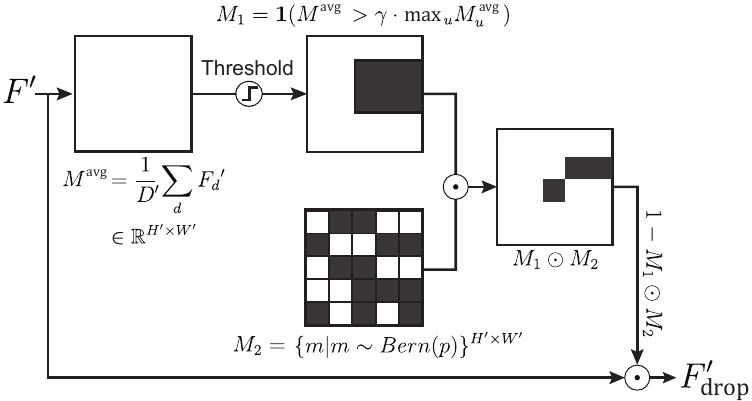}
    \vspace{-1.8em}
    \caption{Dropout mechanism of consistency with attentive dropout}
    \label{fig:dropconsistency}
\end{figure}
Because the value of $\mathcal{F}$ at the most discriminative region is significantly larger than that at the other region, the value of the normalized map $\hat{\mathcal{F}}$ at the less discriminative part but belonging to the object region becomes small.

We introduce consistency with attentive dropout, a method to distribute the activation to the target object region.
We adopt $L_1$ loss between the two feature maps $F$ and $F_\text{drop}$: $F$ is the feedforward result of an intermediate feature map $F'$, and $F_\text{drop}$ is the feedforward result of $F'_\text{drop}$ obtained by intentionally dropping large activations from $F'$.
Fig.~\ref{fig:dropconsistency} shows the overall process of obtaining $F'_\text{drop}$ for consistency with attentive dropout.
In $F'$, the activation at the spatial location whose channel-wise averaged activation is larger than $\gamma$ is dropped with probability $p$. The stochastic dropout prevents all information in the highly activated area from being eliminated. The loss for consistency with attentive dropout is as follows:
\begin{equation}\label{eq:loss_er}
\mathcal{L}_\text{drop} = \|F(x)- F_\text{drop}(x)\|_{1}.
\end{equation}

There have been several attempts that utilize a similar erasing mechanism~\cite{mai2020erasing,choe2019attention,zhang2018adversarial}.
They train a classifier to preserve the predicted labels before and after erasing highly activated features.
In contrast, our method explicitly regularizes a model to yield a similar feature map even after the highly activated features are dropped.
This decreases the dependency on the dropped features, resulting in more evenly distributed activation compared to the other methods.

\subsection{Training Scheme}\label{sec:training_scheme}
With cross-entropy loss for classification, $\mathcal{L}_\text{CE}$, the total cost function is defined as follows:
\begin{equation}\label{eq:loss_tot_step2}
\mathcal{L_\text{total}} = \mathcal{L}_\text{CE} + \lambda_\text{drop} \mathcal{L}_\text{drop} + \lambda_\text{sim} \mathcal{L}_\text{sim} + \lambda_\text{norm} \mathcal{L}_\text{norm},
\end{equation}
where $\lambda_\text{drop}$, $\lambda_\text{sim}$, and $\lambda_\text{norm}$ are hyperparameters for balancing the losses.
The feature direction alignment is better applied after training the classifier to some extent to obtain a suitable feature map for classification.
Thus, for the first few epochs (\ie, the warm stage), we train a model only with $\mathcal{L}_\text{CE}$ and $\mathcal{L}_\text{drop}$:
\vspace{-1pt}
\begin{equation}\label{eq:loss_tot_step1}
\mathcal{L_\text{warm}} = \mathcal{L}_\text{CE} + \lambda_\text{drop} \mathcal{L}_\text{drop}.
\end{equation}
\section{Experiments}

\subsection{Experimental Settings}
\noindent\textbf{Datasets.}
We evaluate our method on two popular benchmarks: CUB-200-2011~\cite{welinder2010caltech} and ImageNet-1K~\cite{russakovsky2015imagenet}.
In the CUB-200-2011 dataset, there are 5,994 images for training and 5,794 for testing from 200 bird species. In the ImageNet-1K, there are approximately 1.3 million images in the training set and 50,000 in the validation set from 1,000 different classes.

\noindent\textbf{Evaluation Metrics.}
Following the work of Russakovsky~\etal~\cite{russakovsky2015imagenet}, we use Top-1 localization accuracy (Top-1 Loc), Top-5 localization accuracy (Top-5 Loc), and localization accuracy with ground-truth class (GT Loc) as our evaluation metrics.
Top-$k$ Loc is the proportion of the images whose predicted bounding box has more than 50\% intersection over union (IoU) with the ground-truth bounding box and whose predicted top-$k$ classes include the ground-truth class.
GT Loc is the localization accuracy with the ground-truth class, which does not consider the classification result.
We also use \texttt{MaxBoxAccV2}~\cite{choe2020evaluation} to evaluate our method.
\texttt{MaxBoxAccV2}$(\delta)$ measures the localization accuracy with ground-truth class with multiple IoU thresholds $\delta\in\{0.3, 0.5, 0.7\}$.

\begin{figure*}[t]
	\centering
    \includegraphics[width=\textwidth]{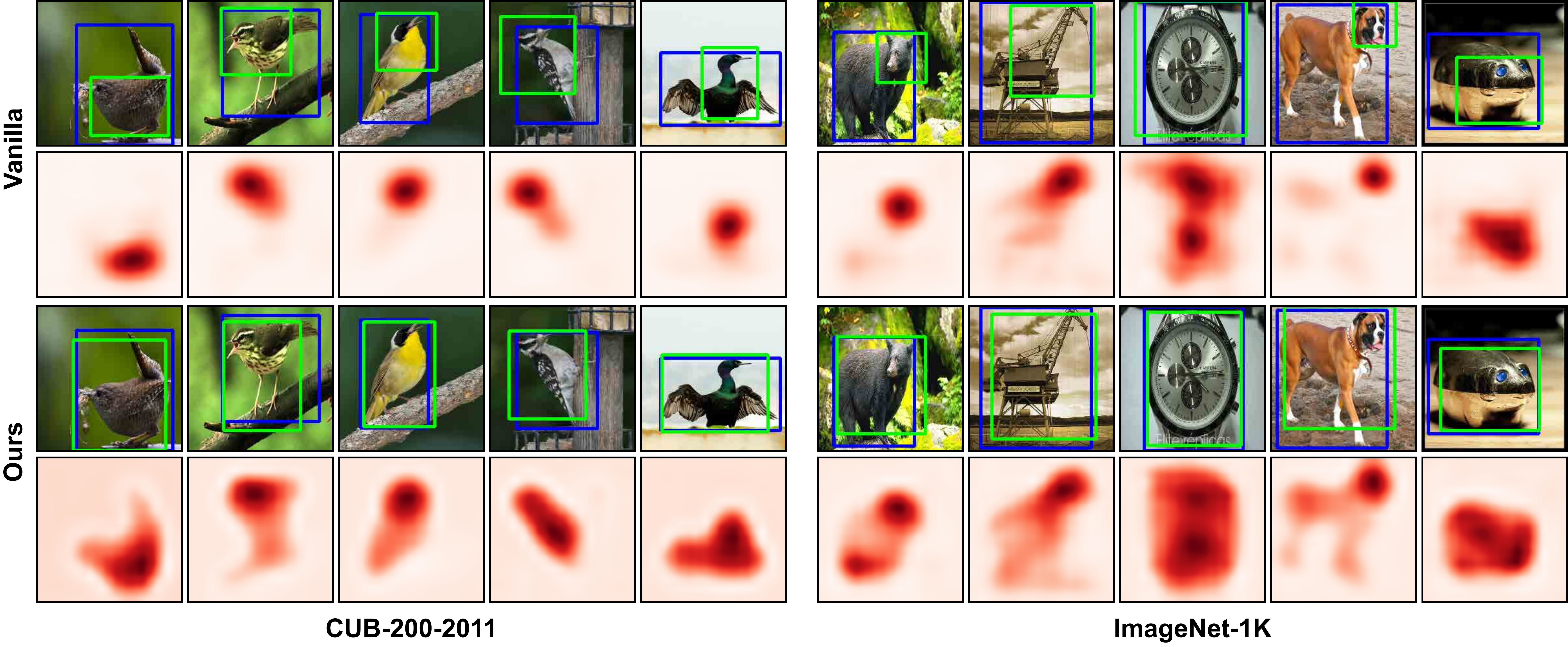}
    \vspace{-2em}
    \caption{Comparison of localization results from the vanilla method and our method on CUB-200-2011 and ImageNet-1K datasets, using VGG16 as a backbone. Blue boxes denote the ground truth bounding boxes and green boxes denote the predicted bounding boxes.}
    \label{fig:compare_cam}
\end{figure*}
\begin{table}[tbp]
\renewcommand{\arraystretch}{0.95}
  \centering
    \begin{tabular}{lccc}
    \Xhline{1pt}\\[-0.95em]
    Method & Top-1 & Top-5 & GT Loc \\
   \hline\hline
\multicolumn{2}{l}{Additional Branch}\\
SLT-Net~\cite{guo2021strengthen}$_{\text{~~CVPR '21}}$ & 67.8 & - & 87.6 \\
ORNet~\cite{xie2021online}$_{\text{~~ICCV '21}}$  &67.74 &80.77 &86.19 \\
FAM~\cite{meng2021foreground}$_{\text{~~ICCV '21}}$  &69.26 &- &89.26 \\
\midrule
\multicolumn{2}{l}{Single Branch}\\
CAM~\cite{zhou2016learning}$_{\text{~~CVPR '16}}$  &44.15 &52.16 &56.00 \\
ADL~\cite{choe2019attention}$_{\text{~~CVPR '19}}$  &52.36 &- & 75.41 \\
DANet~\cite{xue2019danet}$_{\text{~~ICCV '19}}$  &52.52 &61.96 &67.70 \\
EIL~\cite{mai2020erasing}$_{\text{~~CVPR '20}}$  &56.21 &- &- \\
MEIL~\cite{mai2020erasing}$_{\text{~~CVPR '20}}$  &57.46 &- &- \\
DGL~\cite{tan2020dual}$_{\text{~~ACMMM '20}}$ & 56.07 & 68.50	& 74.63 \\
Ki~\etal~\cite{ki2020sample}$_{\text{~~ACCV '20}}$  &57.50 &- &- \\
Bae~\etal~\cite{bae2020rethinking}$_{\text{~~ECCV '20}}$ & 58.96 &  - & 76.30\\
Pan~\etal~\cite{pan2021unveiling}$_{\text{~~CVPR '21}}$  &60.27 &72.45 &77.29 \\
Ours & \textbf{70.83} & \textbf{88.07} & 	\textbf{93.17}\\
    \Xhline{1pt}
    \end{tabular}%
    \vspace{-0.5em}
     \caption{Comparison of localization performance on the CUB-200-2011 test set, based on VGG16.}
  \label{tab:cub_top1loc_vgg}
\end{table}%

\noindent\textbf{Implementation Details.}
We evaluate our method using VGG16~\cite{simonyan2014very} and ResNet50~\cite{he2016deep} as backbone networks.
For VGG16, we adopt the GAP layer following the training settings of the previous work~\cite{zhou2016learning}.
For ResNet50, we set the stride of the third layer to 1.
The attentive dropout is applied before the last pooling layer in VGG16 and after the first block in the fourth layer in ResNet50.
We initialize the networks with the pretrained weights using ImageNet-1K~\cite{russakovsky2015imagenet}. We use a min-max normalization to draw the bounding box from the generated CAM.

\subsection{Comparison with State-of-the-art Methods}
We compare our method to the recent WSOL methods. For other WSOL methods, we report the localization performance of the original papers or that reproduced by \cite{choe2020evaluation,kim2021normalization,bae2020rethinking,tan2020dual}\footnote{https://github.com/clovaai/wsolevaluation}. Our method consistently outperforms existing WSOL methods using a single branch, across the datasets and the backbones by a large margin.

Tab.~\ref{tab:cub_top1loc_vgg} shows the localization performance on the CUB-200-2011~\cite{welinder2010caltech} test set, using VGG16 as a backbone. Our method achieves an 11.87\%p improvement in Top-1 Loc and a 16.87\%p improvement in GT Loc over the work of Bae~\etal~\cite{bae2020rethinking}, which is the state-of-the-art method among the CAM-based methods.
Furthermore, our method outperforms the methods adopting an additional branch for localization. Our method improves Top-1 Loc by 1.57\%p and GT-Loc by 3.91\%p improvement in GT Loc compared to FAM~\cite{meng2021foreground}.

Tab.~\ref{tab:cub_top1loc} shows the results using ResNet50 as a backbone. It shows that our method consistently outperforms the existing methods by a large margin ($>$13\%p), using a different backbone.
\textcolor{black}{Tab.~\ref{tab:imagenet_top1loc} shows the localization performance on the ImageNet-1K~\cite{russakovsky2015imagenet} validation set, based on VGG16 and ResNet50.
Our method achieves the state-of-the-art performance in the ImageNet-1K dataset regardless of the backbone, and only Top-1 Loc with ResNet50 is the second best after I$^2$C with a marginal difference.}

Additionally, we compare our \texttt{MaxBoxAccV2}~\cite{choe2020evaluation} scores with other state-of-the-art methods on the CUB-200-2011 and ImageNet-1K in Tab.~\ref{tab:total_maxbox}.
It shows that our method outperforms the most recent methods by a large margin for all IoU thresholds with various backbones and datasets. Especially, our method improves the score with IoU threshold of 0.7, which is strict accuracy, by 21.0\%p and 17.4\%p with VGG16 and ResNet50 on the CUB-200-2011 dataset, respectively, compared with the work of Ki~\etal~\cite{ki2020sample}.

\begin{table}[tbp]
\renewcommand{\arraystretch}{0.95}
  \centering
    \begin{tabular}{lccc}
    \Xhline{1pt}\\[-0.95em]
    Method & Top-1 & Top-5 & GT Loc \\
   \hline\hline
CAM~\cite{zhou2016learning}$_{\text{~~CVPR '16}}$  & 46.91 &53.57 & - \\
ADL~\cite{choe2019attention}$_{\text{~~CVPR '19}}$  & 57.40 &- & 71.99 \\
CutMix~\cite{yun2019cutmix}$_{\text{~~ICCV '19}}$  &54.81 &- &- \\
DGL~\cite{tan2020dual}$_{\text{~~ACMMM '20}}$ & 60.82 &70.50 & 74.65\\
Ki~\etal~\cite{ki2020sample}$_{\text{~~ACCV '20}}$  &56.10 &- &- \\
Bae~\etal~\cite{bae2020rethinking}$_{\text{~~ECCV '20}}$ &  59.53 &  - & 77.58\\
Ours & \textbf{73.16} &  \textbf{86.68} & \textbf{91.60}\\
    \Xhline{1pt}
    \end{tabular}%
    \vspace{-0.5em}
     \caption{Comparison of localization performance on the CUB-200-2011 test set, based on ResNet50.}
  \label{tab:cub_top1loc}
\end{table}%

Fig.~\ref{fig:compare_cam} shows some examples of localization results from the vanilla method~\cite{zhou2016learning} and from our method on the CUB-200-2011 and ImageNet-1K datasets. It shows that the model trained with our method captures the target object region more accurately than the vanilla model. On the CUB-200-2011 dataset, while the vanilla model fails to identify the tails, legs, and wings of birds, the classifier trained with our method successfully identifies them.

\setuldepth{53.87}
\begin{table}[tbp]
\renewcommand{\arraystretch}{0.95}
  \centering
    \begin{tabular}{lccc}
    \Xhline{1pt}\\[-0.95em]
    Method & Top-1 & Top-5 & GT Loc \\
    \hline\hline
     \multicolumn{2}{l}{Backbone: VGG16}\\
    CAM~\cite{zhou2016learning}$_{\text{~~CVPR '16}}$  & 42.80 &54.86 & - \\
    ACoL~\cite{zhang2018adversarial}$_{\text{~~CVPR '18}}$  & 45.83 &59.43 &62.96 \\
    ADL~\cite{choe2019attention}$_{\text{~~CVPR '19}}$  & 44.92 &- &- \\
    CutMix~\cite{yun2019cutmix}$_{\text{~~ICCV '19}}$  & 43.45 &- &- \\
    I$^{2}$C~\cite{zhang2020inter}$_{\text{~~ECCV '20}}$  & 47.41 &58.51 &63.90 \\
    EIL~\cite{mai2020erasing}$_{\text{~~CVPR '20}}$  &46.27 &- &- \\
    MEIL~\cite{mai2020erasing}$_{\text{~~CVPR '20}}$  &46.81 &- &- \\
    Ki~\etal~\cite{ki2020sample}$_{\text{~~ACCV '20}}$  & 47.20 &- &- \\
    DGL~\cite{tan2020dual}$_{\text{~~ACMMM '20}}$ &   47.66 &58.89 &64.78 \\
    Bae~\etal~\cite{bae2020rethinking}$_{\text{~~ECCV '20}}$ & 44.62 &  - & 60.73\\
    Pan~\etal~\cite{pan2021unveiling}$_{\text{~~CVPR '21}}$  & \ul{49.56} &\ul{61.32} &\ul{65.05} \\
    Ours & \textbf{49.94} & \textbf{63.25} & 	\textbf{68.92}\\
    \midrule
    \multicolumn{2}{l}{Backbone: ResNet50}\\
    ADL~\cite{choe2019attention}$_{\text{~~CVPR '19}}$  & 48.23 & - &61.04 \\
    CutMix~\cite{yun2019cutmix}$_{\text{~~ICCV '19}}$  &47.25 &- &- \\
    Ki~\etal~\cite{ki2020sample}$_{\text{~~ACCV '20}}$  & 48.40 &- &- \\
    Bae~\etal~\cite{bae2020rethinking}$_{\text{~~ECCV '20}}$ &  49.42 & - & 62.20 \\
     I$^{2}$C~\cite{zhang2020inter}$_{\text{~~ECCV '20}}$  & \textbf{54.83} &\ul{64.60} &68.50 \\
    DGL~\cite{tan2020dual}$_{\text{~~ACMMM '20}}$ &  53.41 &62.69 &\ul{69.34} \\
    Ours & \ul{53.76}	& \textbf{65.75}	& \textbf{69.89}\\
    \Xhline{1pt}
    \end{tabular}%
    \vspace{-0.5em}
     \caption{Comparison of localization performance on the ImageNet-1K validation set. The best performance is bold and the second best performance is underlined.}
  \label{tab:imagenet_top1loc}
\end{table}%

\begin{table*}[t]
  \centering
\setlength{\tabcolsep}{3.73pt}
\begin{tabular}{l|cccc|cccc|cccc|cccc}
\Xhline{1pt}
\multirow{4}{*}{Method} &\multicolumn{8}{c|}{CUB-200-2011} &\multicolumn{8}{c}{ImageNet-1K} \\
&\multicolumn{4}{c|}{VGG16} &\multicolumn{4}{c|}{ResNet50} &\multicolumn{4}{c|}{VGG16} &\multicolumn{4}{c}{ResNet50} \\
&\multicolumn{3}{c}{$\delta$} &\multirow{2}{*}{Mean} &\multicolumn{3}{c}{$\delta$} &\multirow{2}{*}{Mean} &\multicolumn{3}{c}{$\delta$} &\multirow{2}{*}{Mean} &\multicolumn{3}{c}{$\delta$} &\multirow{2}{*}{Mean} \\
&0.3 &0.5 &0.7 & &0.3 &0.5 &0.7 & &0.3 &0.5 &0.7 & &0.3 &0.5 &0.7 & \\
\hline\hline
CAM~\cite{zhou2016learning} &96.8 &73.1 &21.2 &63.7 &95.7 &73.3 &19.9 &63.0 &81.0 &62.0 &37.1 &60.0 &83.7 &65.7 &41.6 &63.7 \\
HaS~\cite{singh2017hide} &92.1 &69.9 &29.1 &63.7 &93.1 &72.2 &28.6 &64.6 &80.7 &62.1 &38.9 &60.6 &83.7 &65.2 &41.3 &63.4 \\
SPG~\cite{zhang2018self} &90.5 &61.0 &17.4 &56.3 &92.2 &68.2 &20.8 &60.4 &81.4 &62.0 &36.3 &59.9 &83.9 &65.4 &40.6 &63.3 \\
ADL~\cite{choe2019attention} &97.7 &78.1 &23.0 &66.3 &91.8 &64.8 &18.4 &58.3 &80.8 &60.9 &37.8 &59.9 &83.6 &65.6 &41.8 &63.7 \\
CutMix~\cite{yun2019cutmix} &91.1 &67.3 &28.6 &62.3 &94.3 &71.5 &22.5 &62.8 &80.3 &61.0 &37.1 &59.5 &83.7 &65.2 &41.0 &63.3 \\
Ki~\etal~\cite{ki2020sample} &96.2 &77.2 &26.8 &66.7 &96.2 &72.8 &20.6 &63.2 &81.5 &63.2 &39.4 &61.3 &84.3 &67.6 &43.6 &65.2 \\
HaS + PaS~\cite{bae2020rethinking} &- &- &- &61.2 &- &- &- &61.9 &- &- &- &62.1 &- &- &- &64.6 \\
CALM~\cite{kim2021keep} &- &- &- &64.8 &- &- &- &71.0 &- &- &- &62.8 &- &- &- & 63.4 \\
ADL + IVR~\cite{kim2021normalization} &- &- &- &71.5 &- &- &- &67.1 &- &- &- &63.7 &- &- &- &65.1 \\
Ours & \textbf{99.3} & 	\textbf{93.2} & \textbf {47.8} & \textbf{80.1} & \textbf{99.4} & \textbf{90.4} & \textbf{38.0} & \textbf{75.9} & \textbf{84.8} & \textbf{69.2} & \textbf{45.9} & \textbf{66.6} & \textbf{86.7} & \textbf{71.1}	& \textbf{48.3}	& \textbf{68.7}\\
    \Xhline{1pt}
    \end{tabular}%
    \vspace{-0.5em}
      \caption{Comparison of \texttt{MaxBoxAccV2} scores on the CUB-200-2011 and ImageNet-1K datasets using various backbones.}
  \label{tab:total_maxbox}%
\end{table*}%
\begin{figure}[t]
	\centering
    \includegraphics[width=0.88\columnwidth]{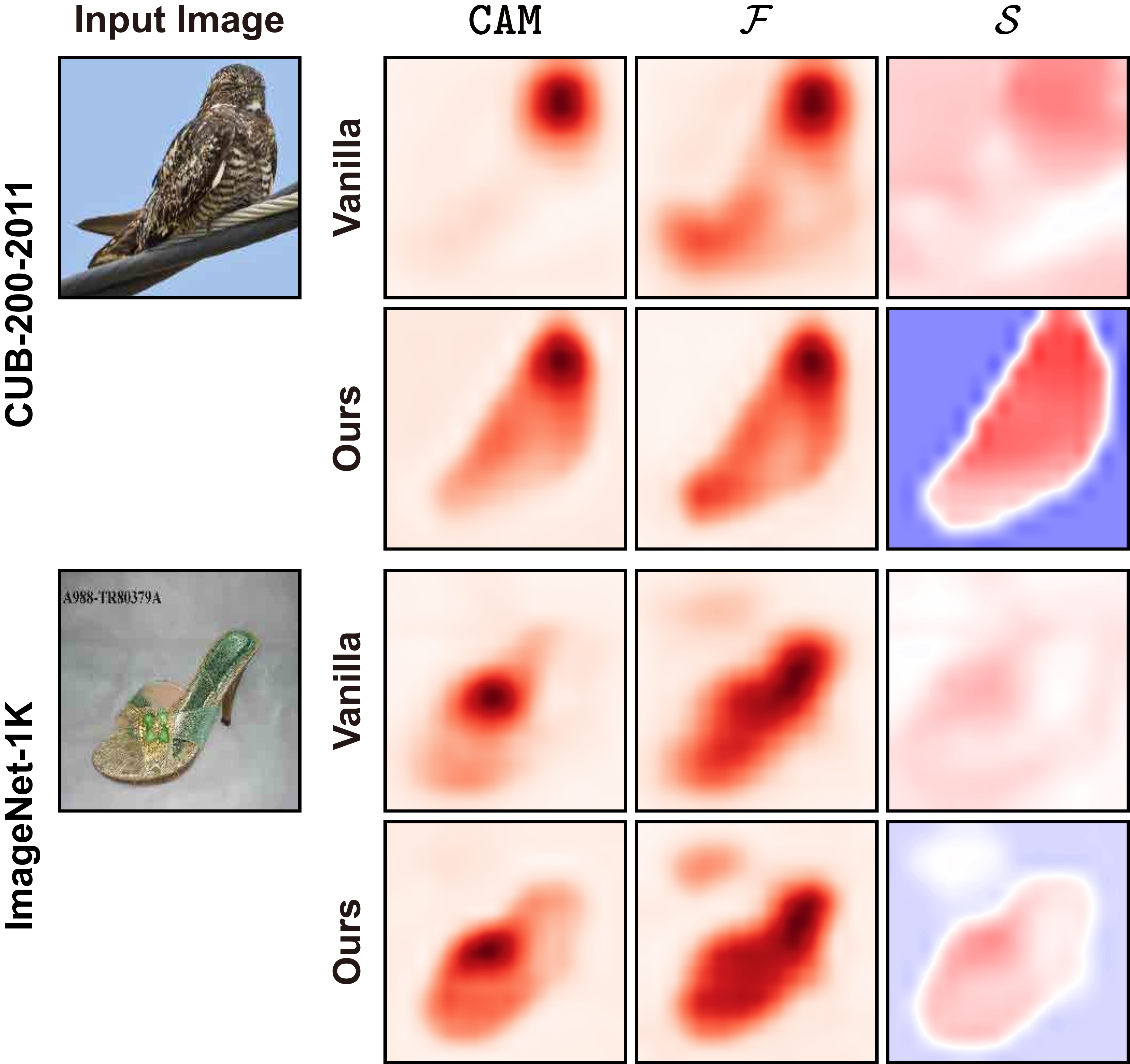}
    \vspace{-0.6em}
    \caption{Comparisons of CAM, $\mathcal{F}$, and $\mathcal{S}$ between the vanilla method and our method on the CUB-200-2011 and ImageNet-1K datasets, using VGG16 as a backbone.}
    \label{fig:compare_sim}
\end{figure}

\subsection{Discussion}
\noindent\textbf{Feature Direction Alignment.}
Through the feature direction alignment, we force $\mathcal{S}$ and $\hat{\mathcal{F}}$ to be high in the object region and to be low in the background region. As Fig.~\ref{fig:compare_sim} shows, the classifier trained with our method yields $\mathcal{S}$ that has a high value in the object region and low value in the background region, different from the vanilla model. It also generates $\hat{\mathcal{F}}$ that has higher activation in less discriminative parts than the vanilla model does.
This makes CAM successfully identify the entire object region. As mentioned in Sec.~\ref{sec:feature_directions}, the feature direction alignment makes $\hat{\mathcal{F}}$ and $\mathcal{S}$ similar, resulting that CAM becomes also similar with them.
We generate a localization map with $\mathcal{F}$ and $\mathcal{S}$ and evaluate the localization performance for each case. We use a min-max normalization when drawing bounding boxes from $\mathcal{F}$. Since negative values in $\mathcal{S}$ denote the background region, we apply a max-normalization on $\mathcal{S}$. Tab.~\ref{tab:perf_sim_norm_cam} shows that the localization results with $\mathcal{F}$ and $\mathcal{S}$ also achieve similar localization performance with CAM. This proves the coincidence between CAM, $\mathcal{F}$, and $\mathcal{S}$ with our method.

\begin{figure}[t]
	\centering
    \includegraphics[width=0.93\columnwidth]{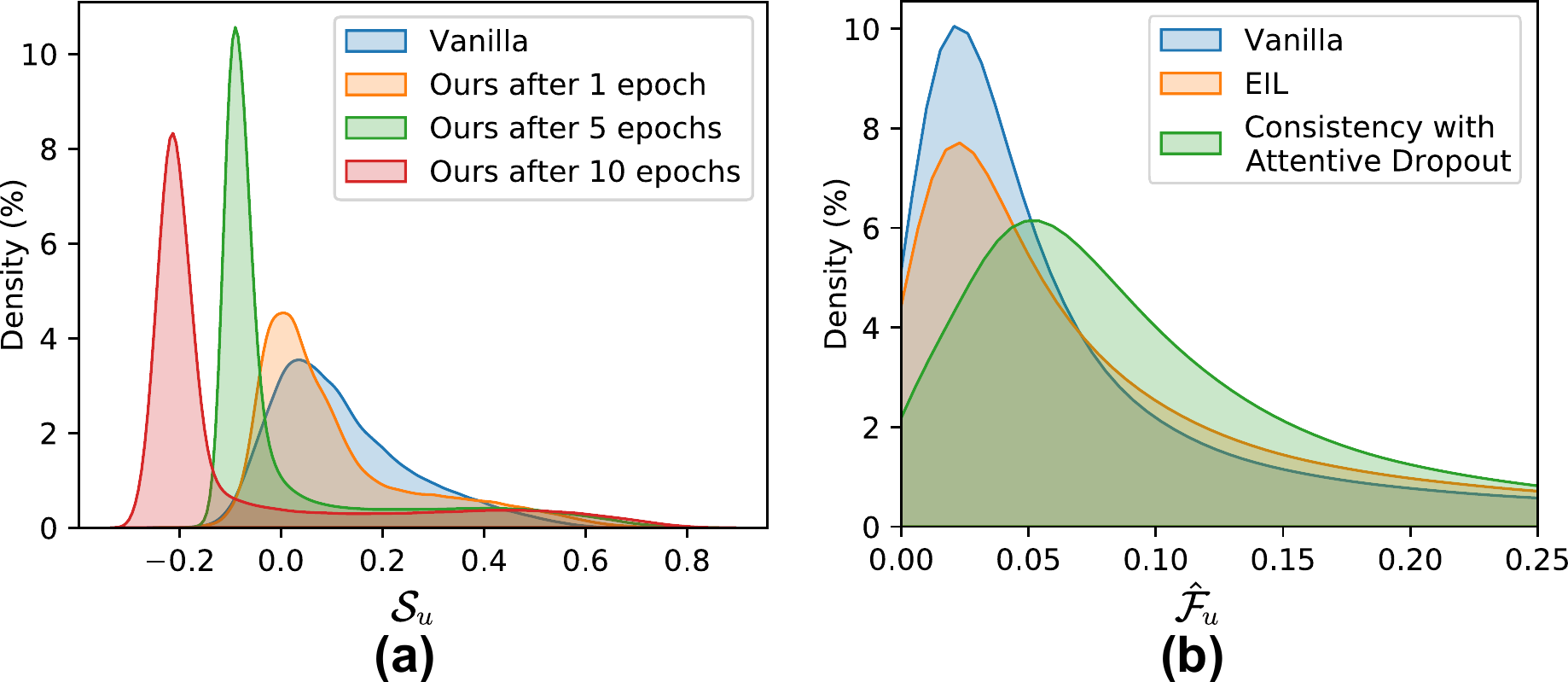}
    \vspace{-0.7em}
    \caption{(a) Comparison of density histogram on $\mathcal{S}_u$ with the vanilla method and our method. (b) Comparison of density histogram on $\hat{\mathcal{F}}_u$ with the vanilla method, EIL, and consistency with attentive dropout. The analyzes are performed on the CUB-200-2011 test set using VGG16 as a backbone.}
    \label{fig:hist}
\end{figure}

Fig.~\ref{fig:hist}(a) shows the distribution of $\mathcal{S}_u$ inside the ground truth bounding boxes from the vanilla method and our method. Note that the bounding boxes include not only the target object but also the background region.
As the training progresses with our method, the similarity gradually splits into negative and large positive values.
This shows that our method effectively increases the similarity for the foreground region and decreases it for the background region.
In contrast, for the vanilla method, the similarity is clustered in small positive values, making no distinction between \mbox{the two}.

\begin{table}[t]
\normalsize
  \centering
  
    \begin{tabular}{cccc}
    \Xhline{1pt}
    Localization map & Top-1  & Top-5 & GT Loc \\
    \hline\hline
     \texttt{CAM}  & 70.83 & 88.07 & 93.17 \\
     $\mathcal{F}$  & 69.90 & 86.68 & 91.96 \\
     $\mathcal{S}$  & 70.38 & 87.64 & 93.13 \\
    \Xhline{1pt}
    \end{tabular}%
    \vspace{-0.5em}
    \caption{Localization performance with various localization maps on the CUB-200-2011 test set, based on VGG16.}
  \label{tab:perf_sim_norm_cam}%
\end{table}%

\noindent\textbf{Consistency with Attentive Dropout.}
Fig.~\ref{fig:hist}(b) compares the effect of our consistency with attentive dropout on the distributions of $\hat{\mathcal{F}}_u$ with the vanilla method and EIL~\cite{mai2020erasing}, the state-of-the-art erasing WSOL method.
Here, the feature direction alignment with $\mathcal{L}_\text{sim}$ and $\mathcal{L}_\text{norm}$ is not applied.
With the vanilla training, most of $\hat{\mathcal{F}}_u$ are very low.
With EIL, overall $\hat{\mathcal{F}}_u$ increase compared with the vanilla method, implying that less discriminative parts become to be highly activated.
With consistency with attentive dropout, the distribution of $\hat{\mathcal{F}}_u$ shifts even more to the right.
This indirectly shows that our proposed method, consistency with attentive dropout, distributes the activation more over the target object region than the other methods. This results that the consistency with attentive dropout achieves higher performance than EIL when used along with feature direction alignment, as shown in Tab.~\ref{tab:compare_eil}. We provide a more detailed analysis in appendix.

\begin{table}[t]
  \centering
    \begin{tabular}{lccc}
    \Xhline{1pt}
    Method & Top-1  & Top-5 & GT Loc \\
    \hline
    \hline
    Align. & 62.27 & 77.48 & 81.93 \\
    EIL~[\textcolor{green}{15}] + Align. & 66.10 & 82.21 & 86.78 \\
    Attentive Dropout + Align. & \textbf{70.83} & \textbf{88.07} & \textbf{93.17} \\
    \Xhline{1pt}
    \end{tabular}%
    \vspace{-0.7em}
     \caption{Comparison of localization performance on the CUB-200-2011 dataset, based on VGG16. Align. denotes the feature direction alignment.}
  \label{tab:compare_eil}%
\end{table}%

\subsection{Ablation Study}\label{sec:ablation}
We perform a series of ablation studies on the CUB-200-2011 dataset using VGG16 as the backbone.

\noindent\textbf{Effect of Each Component.}
Tab.~\ref{tab:ablation} shows the localization performance of the classifier trained with and without each loss term.
Compared to the performance without the proposed loss terms, $\mathcal{L}_\text{drop}$ improves the Top-1 Loc by 7.4\%p and GT Loc by 14.32\%p.
The feature direction alignment using only $\mathcal{L}_\text{sim}$ improves the Top-1 Loc by 9.71\%p and GT Loc by 15.36\%p, which shows the largest improvement among the components.
Adopting $\mathcal{L}_\text{norm}$ improves all metrics more than 5\%p. The feature direction alignment using both $\mathcal{L}_\text{sim}$ and $\mathcal{L}_\text{norm}$ achieves 62.27\% of Top-1 Loc and 81.93\% of GT Loc, which is higher than the performance reported by Pan~\etal~\cite{pan2021unveiling}.
Adoption of all components shows the best performance in all metrics.

\noindent\textbf{Sensitivity to Hyperparameters.}
We analyze the effect of the balancing factors in the loss and the hyperparameters of each loss.

\begin{figure}[t]
	\centering
    \includegraphics[width=0.99\columnwidth]{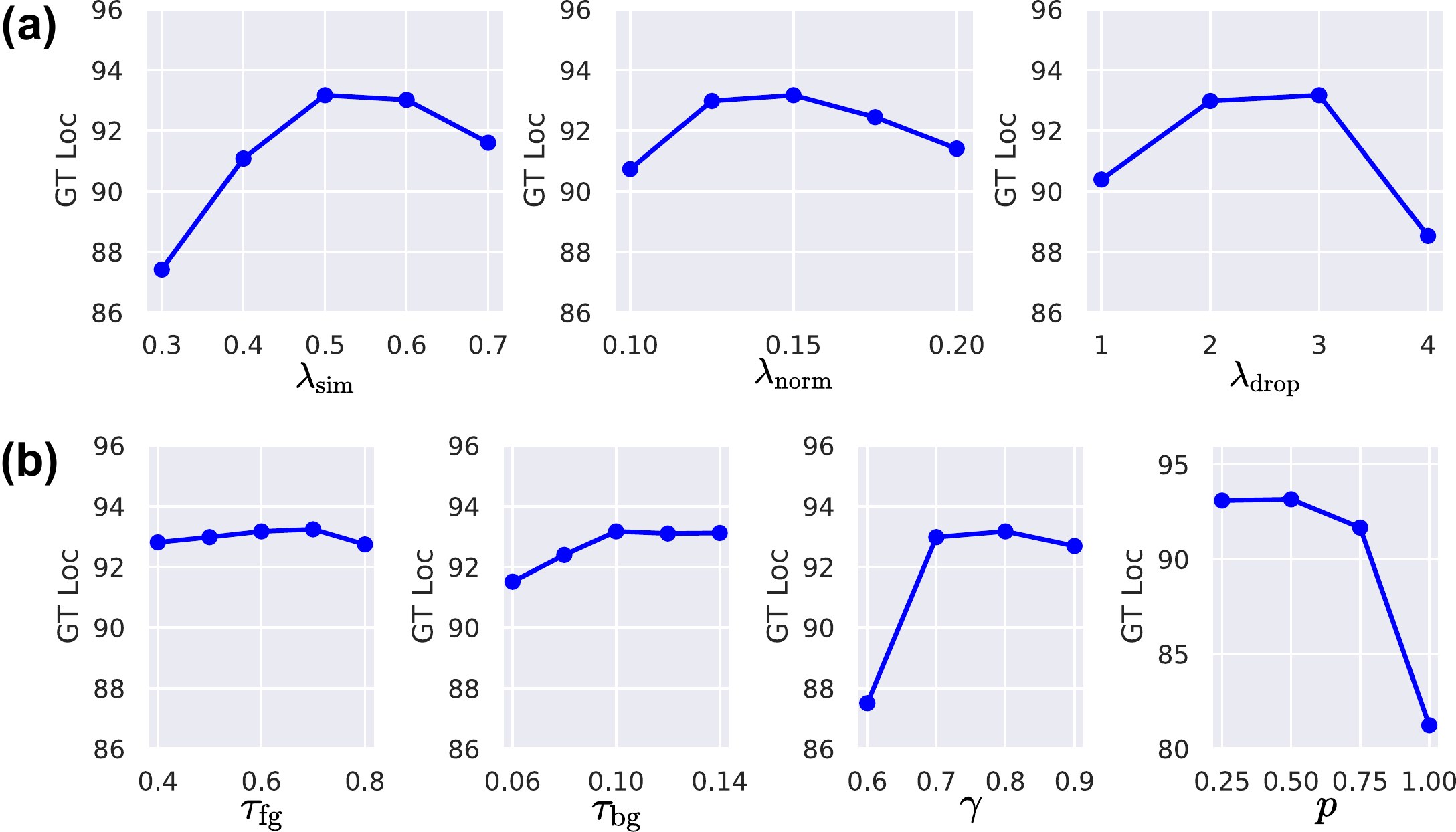}
    \vspace{-0.7em}
    \caption{Effect of (a) balancing factors for loss and (b) various hyperparameters.}
    \label{fig:hyperparams}
\end{figure}

For the balancing factors in loss, we find the best localization performance at 0.5 for $\lambda_\text{sim}$, 0.15 for $\lambda_\text{norm}$, and 3 for $\lambda_\text{drop}$, respectively.
As shown in Fig.~\ref{fig:hyperparams}(a), the localization performance is most sensitively affected by $\lambda_\text{sim}$. $\lambda_\text{norm}$ insignificantly changes the performance.
The performance tends to decrease when the constraint with $\lambda_\text{drop}$ becomes too strong as 4.
\begin{table}[t]
  \centering
    \begin{tabular}{ccc|ccc}
    \Xhline{1pt}
    $\mathcal{L}_\text{drop}$ & $\mathcal{L}_\text{sim}$ & $\mathcal{L}_\text{norm}$ & Top-1  & Top-5 & GT Loc \\
    \hline\hline
    \textcolor{red}{\xmark} &  \textcolor{red}{\xmark} & \textcolor{red}{\xmark} &46.95 &57.23 &60.74 \\
    \textcolor{green}{\cmark}  &  \textcolor{red}{\xmark} &  \textcolor{red}{\xmark}    & 54.35 & 70.37 & 75.06 \\
    \textcolor{red}{\xmark}  &  \textcolor{green}{\cmark} &  \textcolor{red}{\xmark}    & 56.66 & 71.38 & 76.10 \\
    \textcolor{red}{\xmark} &  \textcolor{green}{\cmark}  &  \textcolor{green}{\cmark}& 62.27 &77.48 & 81.93 \\
    \textcolor{green}{\cmark}  &  \textcolor{green}{\cmark} &  \textcolor{red}{\xmark}  & 63.00 & 79.93 & 85.35\\
     \textcolor{green}{\cmark}  &  \textcolor{green}{\cmark} &  \textcolor{green}{\cmark}  & \textbf{70.83} & \textbf{88.07} & \textbf{93.17} \\
    \Xhline{1pt}
    \end{tabular}%
    \vspace{-0.7em}
    \caption{Ablations studies on the CUB-200-2011 test set, based on VGG16.}
  \label{tab:ablation}%
\end{table}%

For the hyperparameters of the feature direction alignment, we set $\tau_\text{fg}$ and $\tau_\text{bg}$ for $\mathcal{L}_\text{sim}$ to 0.6 and 0.1, respectively.
They determine the coarse foreground and background regions.
Fig.~\ref{fig:hyperparams}(b) shows that varying those thresholds has little effect on the performance.
The hyperparameters $\gamma$ and $p$ determine the drop of the activation in the intermediate feature map. $\gamma$ and $p$ for $\mathcal{L}_\text{drop}$ are set to 0.8 and 0.5, respectively.
When $\gamma$ is moderately large between 0.7 and 0.9, there is no significant change in the performance, but when $\gamma$ is too low, \ie, 0.6, the performance decreases.
From the results with various $p$, we observe that stochastic dropout produces little change of GT Loc regardless of the drop probability, but deterministic dropout with a probability of 1.0 yields a significant drop in the localization performance. This indicates that less but sufficient discriminative information should be maintained for a good localization performance.
\section{Conclusion}
In this paper, we find the gap between classification and localization by decomposing CAM from a new perspective. We claim that the misalignment between the feature vector at each location and class-specific weight causes CAM to be activated only in a small discriminative region.
To bridge this gap, we propose a method of aligning feature directions with class-specific weights. We also introduce a strategy to enhance the effect of feature direction alignment.
Extensive experiments demonstrate the effectiveness of the proposed method, which outperforms existing WSOL methods by a large margin.

\noindent\textbf{Limitation.}
There are several hyperparameters to decide in our method. To alleviate the search burden, we discuss a rationale for hyperparameter selection.

\bigskip
\vspace{-1pt}
\noindent\textbf{Acknowledgements:}
This work was supported by Institute of Information \& communications Technology Planning \& Evaluation (IITP) grant funded by the Korea government (MSIT) [NO.2021-0-01343, Artificial Intelligence Graduate School Program (Seoul National University)], LG AI Research, AIRS Company in Hyundai Motor and Kia through HMC/KIA-SNU AI Consortium Fund, and the BK21 FOUR program of the Education and Research Program for Future ICT Pioneers, Seoul National University in 2022.

{\small
\bibliographystyle{ieee_fullname}
\bibliography{egbib}
}

\clearpage
\noindent{\Large \textbf{Appendix}}

\setcounter{section}{0}
\renewcommand\thesection{\Alph{section}}
\setcounter{table}{0}
\renewcommand{\thetable}{A\arabic{table}}
\setcounter{figure}{0}
\renewcommand{\thefigure}{A\arabic{figure}}

\section{Societal Impact}
As deep neural networks require large amounts of data, data-related industries are expanding. One of the prevalent business models of the industries is data annotation. However, the cost of data annotation is burdensome for general users. To reduce the cost, approaches for weakly supervised learning have been proposed, which only requires weaker supervision than fully supervised learning. Since we propose a method of weakly supervised object localization, image-level annotation is sufficient. Our method may threaten the business model of data labeling companies that provide fine-grained labels such as pixel-level annotations and bounding box annotations.

\section{Experimental Details}
We employ SGD optimizer with momentum 0.9 and weight decay $5 \times 10^{-4}$. Following the work of \mbox{Choe~\etal~\cite{choe2020evaluation}}, the networks are divided into two parts, and the learning rate is set differently for those two. VGG16~\cite{simonyan2014very} is divided into old layers and newly added layers when modifying it to VGG16-GAP~\cite{zhou2016learning}, and ResNet50~\cite{he2016deep} is divided into the layers prior to the fourth layer and the others. On the CUB-200-2011 dataset~\cite{welinder2010caltech}, the learning rate is set to $4 \times 10^{-3}$ and $2 \times 10^{-3}$ for the former part of VGG16 and ResNet50, respectively. It is set to $2 \times 10^{-2}$ for the latter part of both backbones. On the ImageNet-1K dataset~\cite{russakovsky2015imagenet}, the learning rate is set to $2 \times 10^{-5}$ and $1 \times 10^{-5}$ for the former part of VGG16 and ResNet50, respectively. It is set to $1 \times 10^{-4}$ for the latter part of both backbones. The proposed method is implemented using PyTorch~\cite{paszke2017automatic}.

Following the work of Choe~\etal~\cite{choe2020evaluation}, we use \texttt{train-fullsup}~\cite{choe2020evaluation} of each dataset as a validation set to select the best model for \texttt{MaxBoxAccV2}~\cite{choe2020evaluation} scores. Please refer to the work of Choe~\etal~\cite{choe2020evaluation} for the details of the dataset.

\section{Additional Results and Discussions}
Additional results and discussions are presented to support the experimental results in the main paper.

\begin{figure*}[t]
	\centering
    \includegraphics[width=\textwidth]{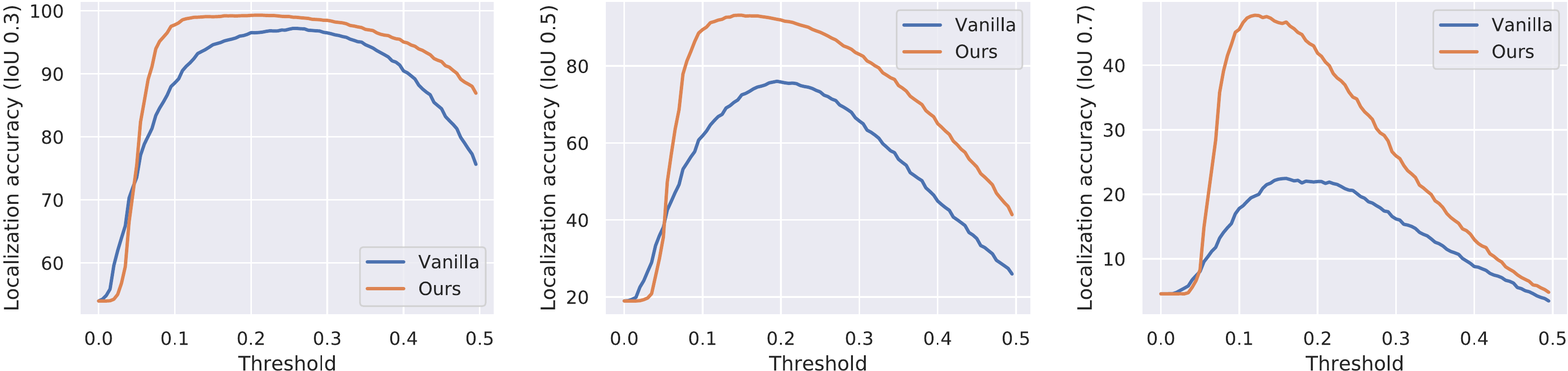}
    \caption{Comparisons of the localization performance when varying threshold on the CUB-200-2011, using VGG16 as a backbone. Three plots show the localization accuracy with IoU 0.3, 0.5, and 0.7. The maximum value of each curve is \texttt{MaxBoxAccV2($\delta$)}.}
    \label{fig:plot_maxbox}
\end{figure*}

\subsection{Sensitivity to Bounding Box Threshold}
A threshold is required to draw a bounding box around an object from a continuous localization map. Fig.~\ref{fig:plot_maxbox} shows that the change of localization accuracy with various IoUs when varying the threshold. In each plot with IoU $\delta$, the maximum value becomes \texttt{MaxBoxAccV2($\delta$)} score. For all $\delta$, the curve of our method is consistently above the curve of the vanilla method, which shows that the superiority of the localization performance of our method does not depend on the threshold. When $\delta$ is 0.3 and 0.5, the curve of our method nearby a maximum value is flatter than the curve of the vanilla method. This shows that our method is less sensitive to the threshold for a bounding box than the vanilla method. When $\delta$ is 0.7, our method is sharper than the vanilla method, but the localization accuracy of our method is more than twice that of the vanilla method, so the flatness comparison is meaningless.

\subsection{Feature Direction Alignment}
\begin{figure*}[ht]
    \centering
    \includegraphics[width=\textwidth]{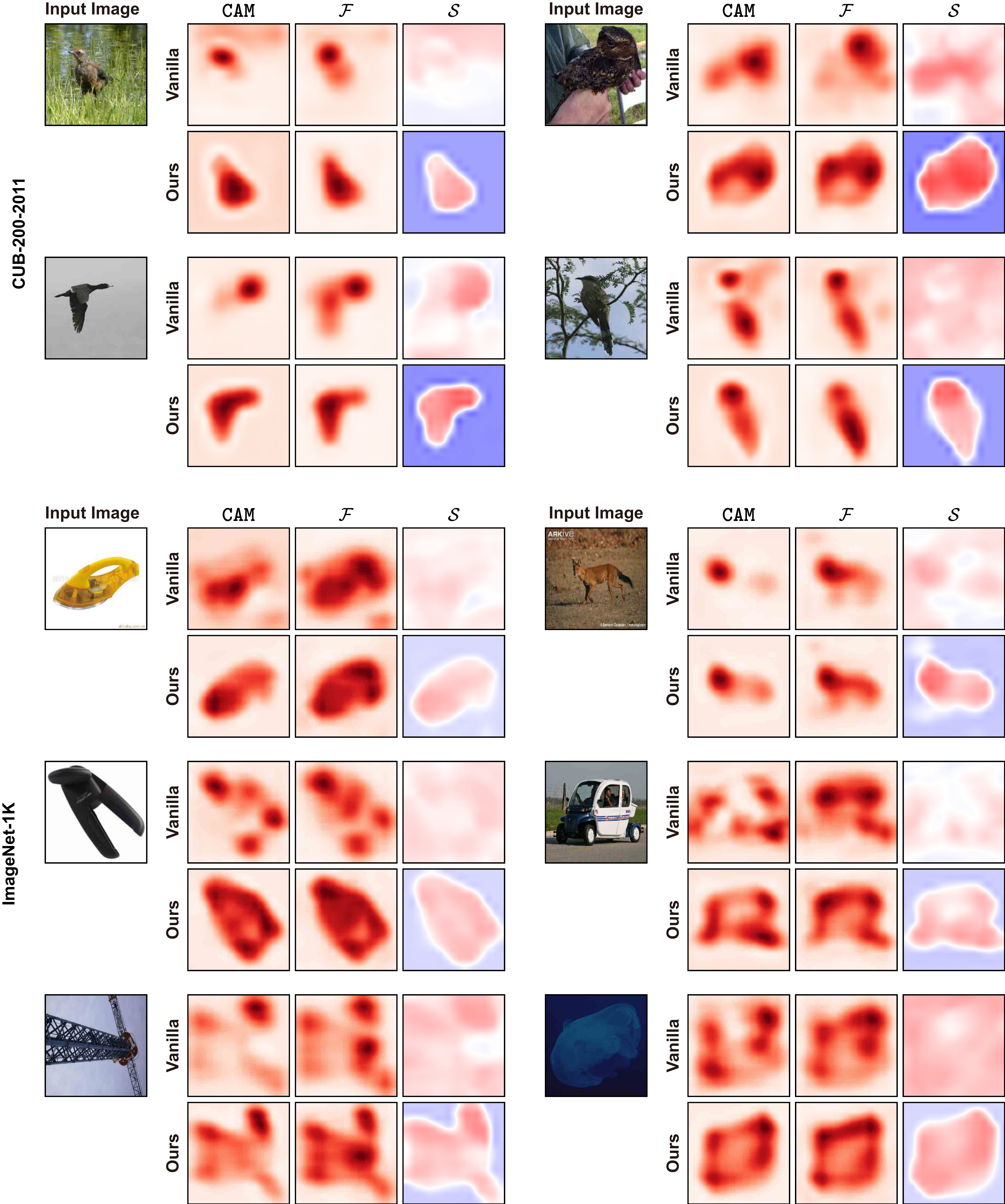}
    \caption{Comparisons of CAM, $\mathcal{F}$, and $\mathcal{S}$ between the vanilla method and our method on the CUB-200-2011 and ImageNet-1K datasets, using VGG16 as a backbone.}
    \label{fig:supp_sim}
\end{figure*}

Fig.~\ref{fig:supp_sim} shows some examples of CAM, $\mathcal{F}$, and $\mathcal{S}$ from the vanilla method and our method on the CUB-200-2011 and ImageNet-1K datasets. In $\mathcal{S}$ from the vanilla method, the overall values are similar and some regions that belong to the object have low values. For instance, in the `car' example (in the second row and second column of the ImageNet-1K dataset), the middle part of the object has low similarity, resulting in low activation in the CAM. Different from the vanilla method, the values of $\mathcal{S}$ from our method are high in the object regions and low in the background regions. Furthermore, the values of $\mathcal{F}$ are high across the entire object region. This makes the CAM that captures more object region.

\subsection{Consistency with Attentive Dropout}
\begin{figure}[t]
	\centering
    \includegraphics[width=0.7\columnwidth]{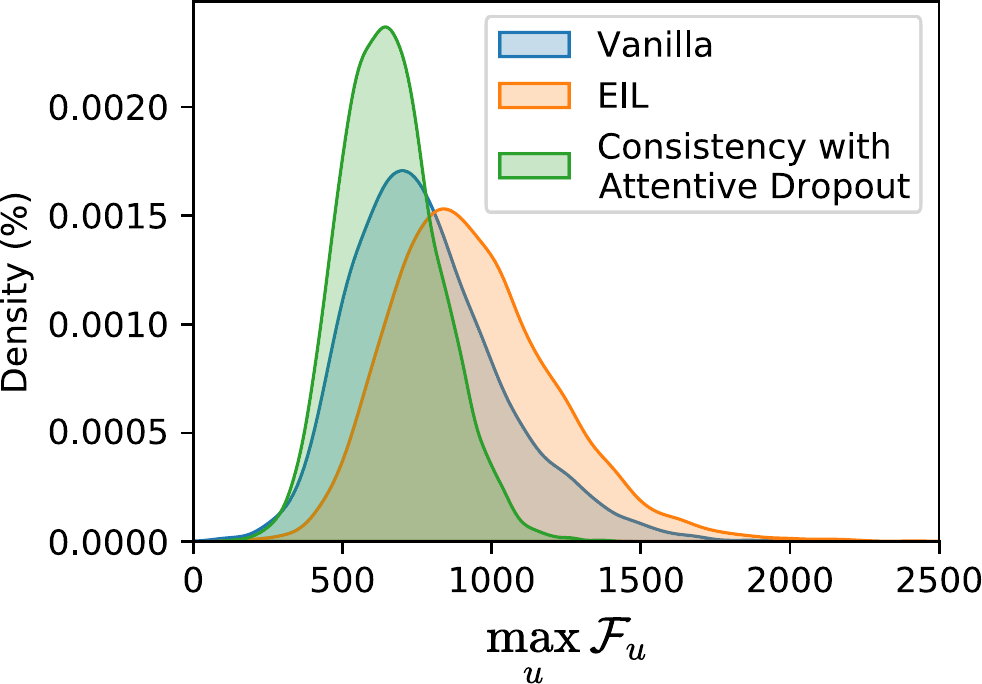}
    \vspace{-0.5em}
    \caption{Comparison of density histogram on $\max_u{\mathcal{F}}_u$ with the vanilla method, EIL, and consistency with attentive dropout. The analyzes are performed on the CUB-200-2011 test set using VGG16 as a backbone.}
    \label{fig:hist_max}
\end{figure}

We compare consistency with attentive dropout with EIL~\cite{mai2020erasing}, the most recent one among the previous erasing methods~\cite{choe2019attention, mai2020erasing,zhang2018adversarial}, on the CUB-200-2011 using VGG16. As mentioned in the main paper, attentive dropout directly regularizes feature activation, whereas EIL indirectly influences the activation through class prediction. Fig.~\ref{fig:hist_max} shows the different effect on feature activation of consistency with attentive dropout and EIL. To encourage a model to predict correct class without highly activated region, EIL enhances the maximum value. In contrast, consistency with attentive dropout reduces the maximum value through regularization. As a result, attentive dropout distributes the activations more effectively than EIL (Fig.~\ref{fig:hist}(b) in the main paper).

\subsection{Localization Results}
\begin{figure*}[ht]
    \centering
    \includegraphics[width=\textwidth]{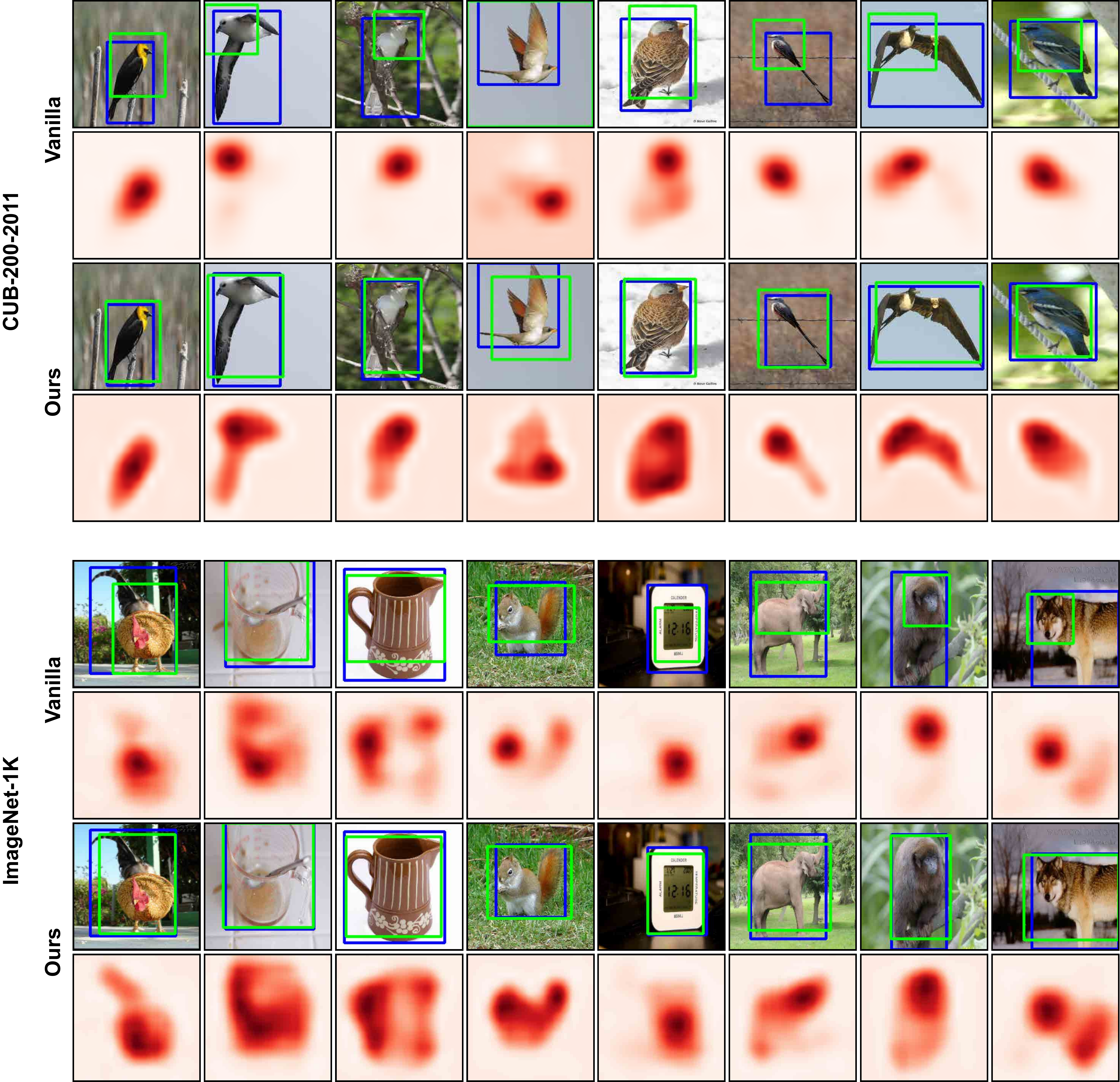}
    \caption{Comparison of localization results from the vanilla method and our method on CUB-200-2011 and ImageNet-1K datasets, using VGG16 as a backbone. Blue boxes denote the ground truth bounding boxes and green boxes denote the predicted bounding boxes.}
    \label{fig:supp_cam}
\end{figure*}

Fig.~\ref{fig:supp_cam} compares the localization results from the vanilla method~\cite{zhou2016learning} and our method on the CUB-200-2011 and ImageNet-1K datasets. While the vanilla method misses the less discriminative parts, \eg, wings and tails of birds and bodies of animals, our method successfully captures the entire object region.

\subsection{Sensitivity to Hyperparameters}
In the main paper, we mention as a limitation that there are several hyperparameters to be decided in our method. We provide further analysis with a different dataset and backbone than that presented in the main paper.

\begin{figure*}[t]
	\centering
    \includegraphics[width=\textwidth]{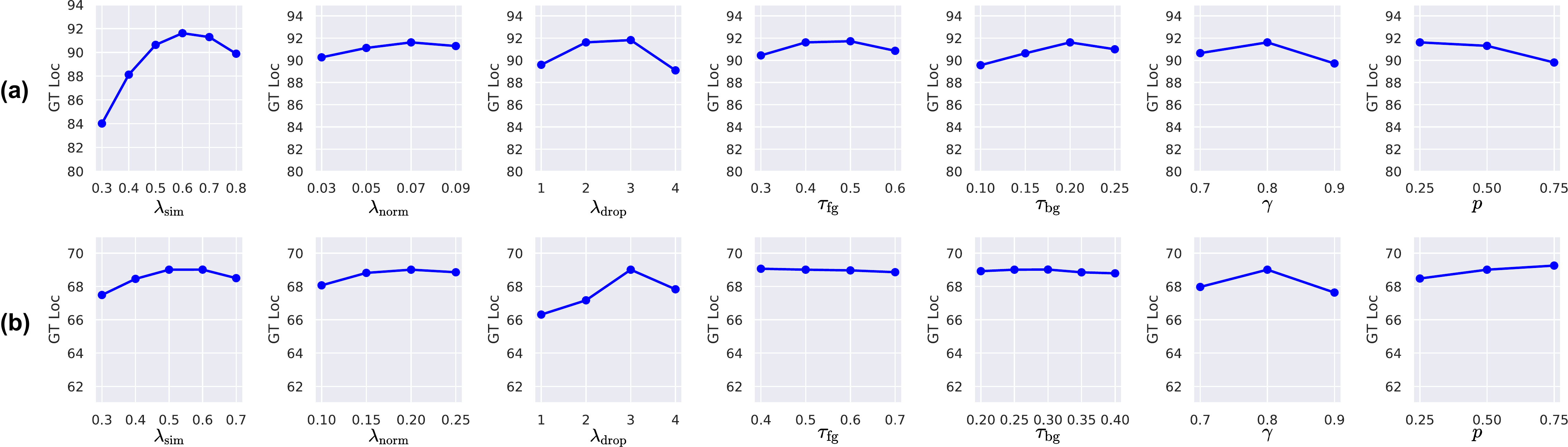}
    \caption{Effect of balancing factors for loss and various hyperparameters. The plots show the results on the CUB-200-2011 test set with ResNet50 and (b) those on the ImageNet-1K validation set with VGG16.}
    \label{fig:supp_plot_hyperparams}
\end{figure*}

\noindent\textbf{CUB-200-2011 on ResNet50.}
We find the best localization performance at 0.6 for $\lambda_\text{sim}$, 0.07 for $\lambda_\text{norm}$, and 2 for $\lambda_\text{drop}$, respectively. The thresholds $\tau_\text{fg}$ and $\tau_\text{bg}$ for $\mathcal{L}_\text{sim}$ are set to 0.4 and 0.2, respectively. The hyperparameters $\gamma$ and $p$ for $\mathcal{L}_\text{drop}$ are set to 0.8 and 0.25, respectively.
Fig.~\ref{fig:supp_plot_hyperparams}(a) shows the change of GT Loc when varying the hyperparameters. The sensitivity to each hyperparameter is similar to that on the CUB-200-2011 dataset using VGG16 as a backbone. $\lambda_\text{sim}$ affects the localization performance the most among hyperparameters.

\noindent\textbf{ImageNet-1K on VGG16.}
The best localization performance is found at 0.5 for $\lambda_\text{sim}$, 0.2 for $\lambda_\text{norm}$, and 3 for $\lambda_\text{drop}$, respectively. The hyperparameters $\tau_\text{fg}$, $\tau_\text{bg}$, $\gamma$, and $p$ are set to 0.5, 0.3, 0.8, and 0.5, respectively.
Fig.~\ref{fig:supp_plot_hyperparams}(b) shows the GT Loc at various hyperparameter values. Different from the CUB-200-2011 dataset, the localization performance is most affected by $\lambda_\text{drop}$. The sensitivities to the other hyperparameters are similar to those with a different dataset and backbone.

\begin{table}[tbp]
  \centering
\begin{tabular}{lc}
    \Xhline{1pt}
    Method & \texttt{PxAP} \\
    \hline
    \hline
    CAM~[\textcolor{green}{28}]$_{\text{~~CVPR '16}}$ & 58.3 \\
    ADL~\cite{choe2019attention}$_{\text{~~CVPR '19}}$ & 58.7 \\  
    CutMix~[\textcolor{green}{28}]$_{\text{~~ICCV '19}}$ & 58.1 \\
    CAM+PaS~[\textcolor{green}{1}]$_{\text{~~ECCV '20}}$ & 59.6 \\   
    ADL+IVR~[\textcolor{green}{9}]$_{\text{~~ICCV '21}}$ & 59.3 \\
    CALM~[\textcolor{green}{10}]$_{\text{~~ICCV '21}}$ & 61.3 \\
    Ours & \textbf{63.7} \\
    \Xhline{1pt}
    \end{tabular}%
    \vspace{-0.5em}
     \caption{Comparison of \texttt{MaxBoxAccV2} scores on the OpenImages dataset using VGG16 as a backbone.}
  \label{tab:openimage_maxbox}
\end{table}%

\noindent\textbf{Discussion.}
The hyperparameters are set differently on the two datasets. This is because their tasks are somewhat different; the classification on the CUB-200-2011 dataset is a fine-grained classification. We additionally evaluate our methods on OpenImages30K~\cite{benenson2019large, choe2020evaluation}, where the task is similar to that on the ImageNet-1K dataset. We use VGG16 as a backbone and set the hyperparameters the same as those on the ImageNet-1K dataset. Note that the OpenImages30K dataset is annotated with a mask, and we use \texttt{PxAP} metric for evaluation, following the work of Choe~\etal~\cite{choe2020evaluation}. As shown in the Tab.~\ref{tab:openimage_maxbox}, our method outperforms the recent methods by a large margin on the OpenImages30K dataset as well, which shows the hyperparameters used on the ImageNet-1K dataset can be applied successfully to a different dataset.

\clearpage

\end{document}